# Robust Real-time Pedestrian Detection in Aerial Imagery on Jetson TX2

Mohamed Afifi*   Yara Ali*   Karim Amer   Mahmoud Shaker   Mohamed ElHelw

Center for Informatics Science
Nile University
Giza, Egypt

`{moh.afifi, y.ali, k.amer, m.serag, melhelw}@nu.edu.eg`

## Abstract

*Detection of pedestrians in aerial imagery captured by drones has many applications including intersection monitoring, patrolling, and surveillance, to name a few. However, the problem is involved due to continuously-changing camera viewpoint and object appearance as well as the need for lightweight algorithms to run on on-board embedded systems. To address this issue, the paper proposes a framework for pedestrian detection in videos based on the YOLO object detection network [6] while having a high throughput of more than 5 FPS on the Jetson TX2 embedded board. The framework exploits deep learning for robust operation and uses a pre-trained model without the need for any additional training which makes it flexible to apply on different setups with minimum amount of tuning. The method achieves ~81 mAP when applied on a sample video from the Embedded Real-Time Inference (ERTI) Challenge where pedestrians are monitored by a UAV.*

## 1. Introduction

Since Krizhevsky et. al. [1] trained a neural network model of multiple convolutional and feedforward layers on large-scale dataset of images for object classification, numerous deep learning architectures have been proposed. One family of these architectures are designed for object detection which entails predicting bounding boxes that enclose objects of interest in a certain image. The current state of the art approaches for this task can be categorized into two categories. The first is based on region proposals such as R-CNN [2], Fast R-CNN [3], Faster R-CNN [4] and Spatial Pyramid Pooling [5]. The second category is for single shot models such as YOLO [6], YOLO v2 [7], YOLO v3 [8] and SSD [9].

In order to deploy the above models onboard of aerial drones, two important aspects have to be taken into consideration. First, typical onboard embedded systems have limited computational power. Second, a sequence of images (*i.e.* video) must be processed. Recent work aimed to address these constraints by creating light-weight versions of original models such as Tiny-YOLO [6] and SSD300 [9]. Other approaches such as Mobile-net [10] and Shuffle-net [11] optimize the base pretrained network to have higher FPS. Lu et. al. [12] incorporated a Long Short Term Memory (LSTM) model to make use of the relation among consecutive frames in a video while Broad et. al. [13] added a convolutional recurrent layer to the SSD architecture to fuse temporal information. Smedt et. al. [14] proposed a framework for on-board real-time pedestrian following on a UAV by merging a constrained search space pedestrian detection algorithm with a particle filter.

This paper proposes an object detection framework that makes use of the temporal information in videos while performing real time inference. The introduced method combines deep learning models pre-trained on large scale dataset of single images but with different input resolutions to perform real-time pedestrian detection using Nvidia Jetson TX2 embedded computing board.

## 2. The Proposed Framework

The proposed framework starts by running YOLO-v3 [8] object detection network with input size of 416x416 that detects pedestrians in the entire image. However, YOLO does not take into consideration the temporal relation between consecutive frames and often detects an object in a number of frames then drops the same object in the following frames. It also fails to localize pedestrians accurately in frames that have high density of and occlusion between pedestrians. To overcome these two problems, we propose crop positions in the next frame based on locations that contain objects in the current frame. Boxes that are close to each other are merged together to a single crop. By running inference only at locations that contain objects, we avoid applying computations on vacant parts of the image.

Proposed crops are subsequently passed on to YOLO for object detection. For each crop, YOLO outputs bounding boxes with high and low confidence scores. The former are considered correct detections and are added to a set of detected objects whereas the latter boxes are only considered if they overlap with boxes correctly detected in the previous frame. To take into account new pedestrians that appear in the scene which may not be captured by the crops, we run YOLO on the whole image every 5 frames and update our crops. It was noticed that in cases where the scene includes many detected pedestrians, the framework fails to achieve the required 5 FPS. In this case, we switch to faster inference mode by monitoring the FPS frame by frame and only run YOLO 416 if the FPS is about to drop.

## 3. Experiments and Results

The proposed framework is tested on the sample video provided by the ERTI Challenge. The video was annotated using a publicly available tool called DarkLabel [15] to evaluate our work quantitatively and tune system parameters. The framework is tested with different input sizes where it achieved good performance and better qualitative and quantitative results than YOLO, due to our framework good performance in scenes with high occlusions and many pedestrians, as in figure 1. For input size 608x608, YOLO achieved 3 FPS and 73.9 mAP score whereas our framework achieved 4.2 FPS and 81 mAP. For input size 416x416, YOLO attained 64 mAP score with execution speed of 6.6 FPS while our framework attained 5.4 FPS and ~81 mAP. It can be seen that our framework is more robust to input size variations since inference only runs on crops taken at sparse locations in the input frame and we can decrease the input size and get higher FPS without losing much accuracy.

## 4. Conclusions and Future Work

This work provides a framework for detecting pedestrians in videos captured by hovering unmanned aerial drones. The framework applies the YOLO object detection network on proposed image regions, instead of the entire image, resulting in higher detection accuracy while sustaining framerates above 5 FPS. Future work will include incorporating the proposed regions within the YOLO network architecture and utilizing structural constraints [16] to further enhance detection results.

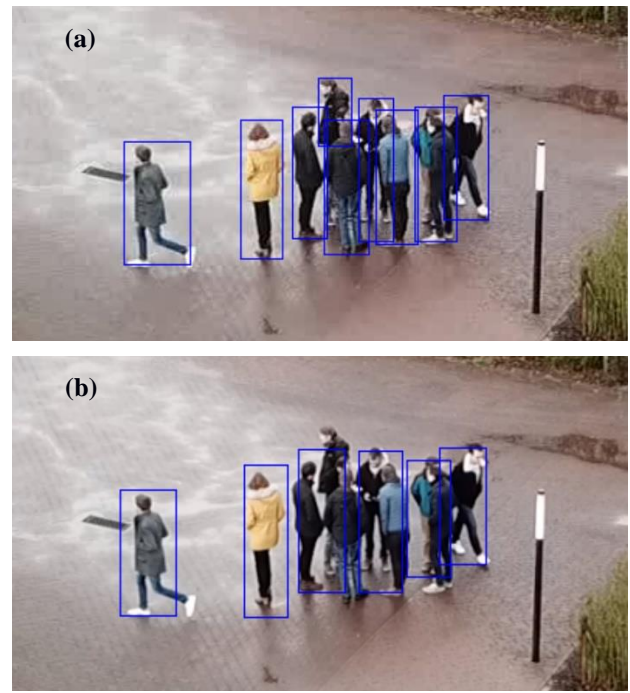

**Figure 1**: The proposed framework performs better in scenes with high occlusions and many pedestrians compared to YOLO-v3. Shown above are framework output (a) and YOLO 416 output (b).